\documentclass{article}
\usepackage[preprint]{nips_2018}      

\usepackage[linesnumbered,ruled]{algorithm2e}
\usepackage{natbib}
\usepackage[T1]{fontenc}    
\usepackage{hyperref}       
\usepackage{url}            
\usepackage{booktabs}       
\usepackage{amsfonts}       
\usepackage{nicefrac}       
\usepackage{microtype}      
\usepackage{amsmath}
\usepackage{amssymb}
\usepackage{amsthm}
\usepackage{graphicx}
\usepackage{physics}
\usepackage{array}
\usepackage{multirow}
\usepackage{subcaption}
\usepackage{wrapfig}
\usepackage{mathtools}
\usepackage{caption}
\usepackage{makecell}
\usepackage{stmaryrd}
\usepackage{wrapfig}
\usepackage{xcolor}
\DeclareMathOperator{\argmax}{\arg\max}

\newcommand{\method}{{AbdGen}}

\newcommand{\citesec}[1]{Section \ref{#1}}

\newcommand{\citeeq}[1]{Equation \ref{#1}}

\newcommand{\citefig}[1]{Figure \ref{#1}}
\newcommand{\citetab}[1]{Table \ref{#1}}

\title{Pre-Training Meta-Rule Selection Policy for Visual Generative Abductive Learning}

{
  \author{Yu Jin$^1$\quad Jingming Liu$^1$ \quad Zhexu Luo$^1$ \quad Yifei Peng$^1$ \quad Ziang Qin$^1$ \\ \textbf{Wang-Zhou Dai}$^2$ \quad \textbf{Yao-Xiang Ding}$^1$\thanks{Corresponding author.} \quad \textbf{Kun Zhou}$^1$ \\[1.25ex]
$^1$State Key Laboratory of CAD\&CG, Zhejiang University \\
$^2$National Key Laboratory for Novel Software Technology, Nanjing University \\[1.25ex]
\texttt{\{jinyu99907,jml754457,zhexuluo,yifeidmw,qinziang19937\}@gmail.com}\\
\texttt{dingyx.gm@gmail.com,daiwz@lamda.nju.edu.cn,kunzhou@acm.org}
}}

\begin{document}
\maketitle              
\begin{abstract}
  Visual generative abductive learning studies jointly training symbol-grounded neural visual generator and inducing logic rules from data, such that after learning, the visual generation process is guided by the induced logic rules. A major challenge for this task is to reduce the time cost of logic abduction during learning, an essential step when the logic symbol set is large and the logic rule to induce is complicated. To address this challenge, we propose a pre-training method for obtaining meta-rule selection policy for the recently proposed visual generative learning approach AbdGen~\citep{peng2023generating}, aiming at significantly reducing the candidate meta-rule set and pruning the search space. The selection model is built based on the embedding representation of both symbol grounding of cases and meta-rules, which can be effectively integrated with both neural model and logic reasoning system. The pre-training process is done on pure symbol data, not involving symbol grounding learning of raw visual inputs, making the entire learning process low-cost. An additional interesting observation is that the selection policy can rectify symbol grounding errors unseen during pre-training, which is resulted from the memorization ability of attention mechanism and the relative stability of symbolic patterns. Experimental results show that our method is able to effectively address the meta-rule selection problem for visual abduction, boosting the efficiency of visual generative abductive learning. Code is available at \url{https://github.com/future-item/metarule-select}.
\end{abstract}
  
\section{Introduction}
\label{sec:intro}
Building neuro-symbolic visual generation model with logic symbol groundings, where the visual generative process is guided by logic rules, has been an emerging research topic recently~\citep{misino2022vael,peng2023generating}. The symbol-grounded neural visual generator is appearing not only because the generation results are interpretable and stable, but also because they lead to strong rule-based generalization ability and capability of inducing logical generative rules from data. Among the initial researches, AbdGen~\citep{peng2023generating} promotes the idea of abductive learning~\citep{zhou2019abductive,10.5555/3454287.3454540} into visual generative learning, enjoying the advantage of doing neural visual generator learning, symbol grounding, and logic rule induction simultaneously from dataset with limited labeling information for symbol grounding.

For further enabling AbdGen to handle large sets of logical symbols and complicated logic generation rules, which is common in visual generation tasks with complicated spacial and temporal dependencies, a major obstacle is the time cost of logical abduction during training. The logic abduction process of AbdGen is done by Prolog-based logic programming system, which is non-differentiable. Even though it is advantageous for its stability and interpretability, the efficiency is an issue especially compared with the neural module, which is differentiable and computationally efficient. The logical abduction process of AbdGen follows the framework proposed in MetaAbd~\citep{dai2020abductive}, which involves both symbol grounding and rule induction. The learning process integrates symbol grounding abduction, and meta-interpretative learning~\citep{muggleton2017meta}, such that the induction of the rule is transformed into the search over the possible rule space induced by the meta-rule set. The size of the meta-rule set significantly affects the time cost of rule induction. The smaller the size, the more efficient the abduction process. On the other hand, for supporting inducing general logic rules without too much tweaking and hand-tuning, a general and larger-sized meta-rule set is preferrable. This trade-off is the main technical challenge to deal with for our work.

In this paper, we propose a pre-training strategy for learning the meta-rule selection policy aiming at boosting the efficiency of logical abduction during AbdGen training. The pre-trained selection policy is the key to deal with the trade-off between generality and efficiency for meta-rule set. For AbdGen training, a general meta-rule set with complete meta-rules can be kept. Given a training instance which includes positive and negative cases of images without symbol grounding labels, the meta-rule selection policy chooses a small set of meta-rules for inputting into the logical reasoning module, hence boosting the abduction efficiency. Utilizing the embedding representation of symbol groundings in AbdGen model, our selection model further builds embedding representations of meta-rules, and utilizes the attention module for meta-rule selection, which can be effectively integrated with both neural model and logic reasoning system. Furthermore, benefited from the disentanglement of symbol and sub-symbol representations of AbdGen, the meta-rule selection policy can be pre-trained under pure symbolic data, which does not involve symbol grounding learning process, thus is of low time cost. Finally, a potentially technical issue for the pre-trained selection policy is that unseen symbol grounding errors can appear during the AbdGen training stage, especially at the beginning of training. While it is interesting to observe that the pre-trained selection policy has relatively strong toleration ability for the errors. This verifies the memorization ability of attention mechanism as well as the relative stability of symbolic patterns. Experimental results show that our method can be utilized to significantly reduce the time cost of abduction, even when the abductive learning task is unseen during the pre-training stage.

\section{Related Work}
\label{sec:rw}
\subsection{Neuro-Symbolic Visual Generative Learning}
Visual generative learning has received great attention in recent years, in particular multi-modal generation such as text-to-vision generation~\citep{ramesh2021zero,rombach2022high,saharia2022photorealistic}. Despite the great success, the generation results suffer from hallucination, still lacking strong controllability, interpretability, and stability. One promising solution is to augment the generation process with symbolic modules to utilize their reasoning power instead of solely relying on black-box neural models.

One possible solution is fully end-to-end differentiable neuro-symbolic models, which is still not quite common for generative models, but is common for addressing inference tasks such as visual question answering~\citep{yi2018neural,hsu2024s}. Such models are relatively efficient due to the fully differentiable properly. On the other hand, the neural part usually needs specific designs to achieve particular functionality. In comparison, traditional logic reasoning models usually enjoys the benefit of generality, while the efficiency is usually a major issue to utilize them in complicated tasks. While in our view, this issue can be addressed by guiding the search-based reasoning process based on the data-driven neural model. The method proposed in this work pursuits this idea.

Building neuro-symbolic visual generative model is still an emerging research topic, in special logic-based models. One pioneer work is VAEL~\citep{misino2022vael}, which integrates the DeepProblog~\citep{manhaeve2018deepproblog} framework with autoencoder visual generators, achieving impressive ability for visual objects generation based on logic rules, as well as strong generalization when the training and testing logic rules differ. On the other hand, AbdGen~\citep{peng2023generating} is based on the visual generative abductive learning framework, which enables both symbol grounding and rule induction, two central tasks in neuro-symbolic visual generation. Our work is built on AbdGen due to its strong disentanglement ability of symbolic and sub-symbolic representations and its effective and general logic induction functionality. 

\subsection{Meta-Rule Selection}
Designing proper inductive bias is known as a grand challenge for inductive logic programming (ILP)~\citep{cropper2022inductive}, which is also the situation for visual generative abductive learning, as a neuro-symbolic extension to the ILP problem. One of the major formulations of the inductive bias is the choice of the meta-rules~\citep{emde1983discovery,de1992interactive,kietz1992controlling,muggleton2017meta}.
Meta-rules are second-order rules which define the structure of learnable programs, whose choice significantly affects the learning performance. As discussed in \citesec{sec:intro}, there is a trade-off for universality and efficiency in deciding the size of the meta-rule set. Even though a number of pioneer researches have been conducted on identifying the suitable meta-rule set~\citep{cropper2015logical,tourret2019sld,cropper2020logical}, this problem is still under-explored and remain unaddressed. Our work builds upon the idea of learning meta-rule selection model from data, which is also studied in previous researches~\citep{mccreath1995extraction,ferilli2004automatic,picado2017towards} under pure symbolic learning tasks. While our method faces neuro-symbolic learning scenario, and builds upon the benefits of both neural models and symbol modules, with no previous researches in this direction.

\section{Problem Setup}
\label{sec:ps}
\subsection{Visual Generative Abductive Learning}


\noindent{\bf Symbol-grounded conditional generation.} We first formalize the neuro-symbolic visual generation task for our learned model to address. An example is illustrated in \citefig{fig:condgen}. The generation process takes one or more input images $x$ as conditions. The input images have their logical symbol groundings $z_{ground}$ and sub-symbolic style $z_{sub}$. The generation is controlled by a set of background logic rules $B$, and the output is a new set of images $\hat x$ with their corresponding symbol groundings $\hat z_{ground}$ and $\hat z_{sub}$, satisfying following constraints:
\begin{equation}
B \cup z_{ground} \vDash \hat z_{ground}, \; \hat z_{sub} = z_{sub},
\label{eq:induce}
\end{equation}
where $\vDash$ denotes logical entailment, ensuring that the symbol groundings $\hat z_{ground}$ of the generated images are consistent with the logic rule $R$ and the symbol grounding $z_{ground}$ of the input image. Furthermore, the generated images maintain the sub-symbolic style $z_{sub}$ of the input.

\begin{figure}[t]
\centering
\includegraphics[width=0.7\textwidth]{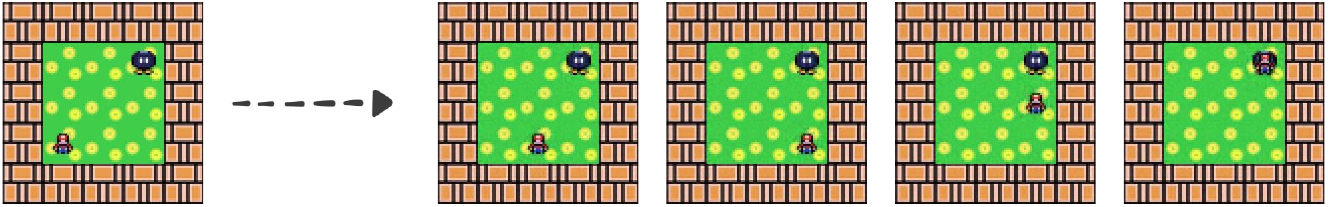}
\centering
\caption{Example of symbol-grounded conditional generation. Given the first picture and the symbolic rule {\it Mario moves with right priority followed by up priority towards the target}, a sequence of images would be generated. Here the symbol to be grounded for each image is the position of Mario, and the sub-symbolic style includes all visual features unchanged among images.} 
\label{fig:condgen}
\end{figure}

{\noindent\bf Neuro-symbolic visual generator.} The generation model builds upon the neuro-symbolic visual generator proposed in AbdGen~\citep{peng2023generating}, which consists of both neural and symbolic parts. The neural part $(E_{sym}, E_{sub}, V, G)$ consists of a symbolic encoder $E_{sym}$, a sub-symbolic encoder $E_{sub}$, a symbol grounding module $V$, and a decoder $G$. The encoders $E_{sym}$ and $E_{sub}$ are responsible for transforming input image $x$ into latent symbolic representation $z_{sym}$, which preserves symbolic information {\it but remains ungrounded}, and sub-symbolic style representation $z_{sub}$. The symbol grounding module $V$, which builds upon vector-quantized structure~\citep{van2017neural}, is responsible for grounding $z_{sym}$ into grounded representations $z_{ground}$. All the representations are assumed to be vector embeddings. This encoding process can be expressed with 
\begin{equation}
  z_{sym} = E_{sym}(x),\; z_{ground} = V(z_{sym}), \; z_{sub} = E_{sub}(x).\ 
\label{eq:gen}
\end{equation}
Once $z_{ground}$ and $z_{sub}$ are identified, they are provided into the logical reasoning system, which is {\it non-differentiable and Prolog-based}, with the background knowledge $B$ inside. Then $\hat z_{ground}$ and $\hat z_{sub}$ are identified based on \citeeq{eq:induce}. Given $\hat z_{ground}$ and $\hat z_{sub}$ as inputs, the decoder $G$ generates the output image $\hat x$:
$G(\hat z_{ground} \oplus \hat z_{sub}) = \hat x$, 
where $\oplus$ denotes the embedding concatenation. The above model structure reveals the essentialness of building embedding representations of symbol groundings, as well as the disentanglement between symbolic and sub-symbolic information, which are guaranteed by on the training mechanism proposed in AbdGen.

{\noindent\bf Visual generative abductive learning}: The learning process for AbdGen involves both training the above symbol-grounded neuro-symbolic visual generator and inducing the background knowledge $B$ from data. Following classical ILP learning scenario, the data is assumed to include both positive and negative cases of raw images, which are generated w.r.t. the hidden background knowledge $B$. Furthermore, we assume that the symbol groundings of only a very small subset of images are labeled, which are denoted as $Z_{ground}^L$. Furthermore, only a subset of $B$, which is denoted as $B_L$, is given to the learner. The learning problem can formulated into
\begin{equation}
\argmax_{\theta, Z^U_{ground}, B_U} \mathcal{L}(X, Z^L_{ground}, B_L | \theta, Z^U_{ground}, B_U),
\label{eq:rulelearn}
\end{equation}
where $\mathcal L$ is the data likelihood, $\theta$ is the parameters of the neuro-visual generator, $Z^U_{ground}$ is the unlabeled symbol groundings in the dataset, and $B_U$ is the missing logic rules in $B$ to induce. This learning problem is solved by the contrastive meta-abduction method proposed in~\citet{peng2023generating}, which pursuits meta-interpretative and meta-abductive learning~\citep{muggleton2017meta,dai2020abductive} to visual generative abductive learning. We omit the details of this method due to space limitations. The key is that the method iteratively generates {\it pseudo groundings} $z^{pseudo}_{ground}$ for the unlabeled cases, which are gradually rectified during the learning process, and conducts logic induction to obtain $B_U$ based on $z^{pseudo}_{ground}$ and a set of candidate meta-rules $M \subset B_L$ in the given background knowledge. The target of our work is to pre-train a meta-rule selection policy to obtain a shrinked set of meta-rules adaptively w.r.t. different $z^{pseudo}_{ground}$, instead of using a fixed set of $M$ throughout the learning process.
\subsection{Meta-Rule Selection Policy}
\label{sec:Policy}
We provide the concrete formulation of the meta-rule selection policy. During the training process of AbdGen, each unlabeled training instance consists of a fixed number of positive and negative cases, whose pseudo-groundings are prediced by the model and denoted $z^{pseudo}_{ground}$ as above. Furthermore, a {\it meta-rule pool} $M$ is prepared, which consists of all candidate meta-rules for learning. The meta-rule selection policy $\pi(z^{pseudo}_{ground}): \mathcal Z \mapsto 2^M$ is a mapping from the symbol grounding space $\mathcal Z$ to the product space of the meta-rule pool, for selecting a proper subset $\mathtt m \subset M$ of meta-rules for $z^{pseudo}_{ground}$. The rules in $\mathtt m$ are considered as the most relevant to the current pseudo groundings. The meta-rule selection policy is obtained from a pre-training process as introduced below.

{\noindent\bf Selection policy pre-training.} For obtaining the selection policy $\pi$, during the pre-training stage, we assume there exist a pure symbolic dataset, which is generated from a set of background knowledge sets $\mathbf B = \{B'_1, B'_2, \dots, B'_K\}$. Each of the instances in the dataset is generated by first sampling a background knowledge set $B'$ from $\mathbf B$, and then sampling positive and negative cases with their symbol groundings $z_{ground}$. Different from the visual generative abductive learning in AbdGen, 1) the instances consist of only symbol groundings without raw image inputs; 2) the groundings are fully labeled. Such dataset is easier to be collected than raw image data, or even can be directly synthesized in applications. Furthermore, each $B'\in\mathbf B$ indicates a different latent generation rule. The learner is given the same meta-rule pool $M$, and targeting at obtaining the selection policy $\pi$ from the dataset. The learned policy is expected to be applied on the AbdGen learning task where the background knowledge set $B$ is outside of $\mathbf B$. In the experiments, we verify that our approach indeed achieves this ability.       



\begin{figure*}[t]
\centering
\includegraphics[width=.9\linewidth]{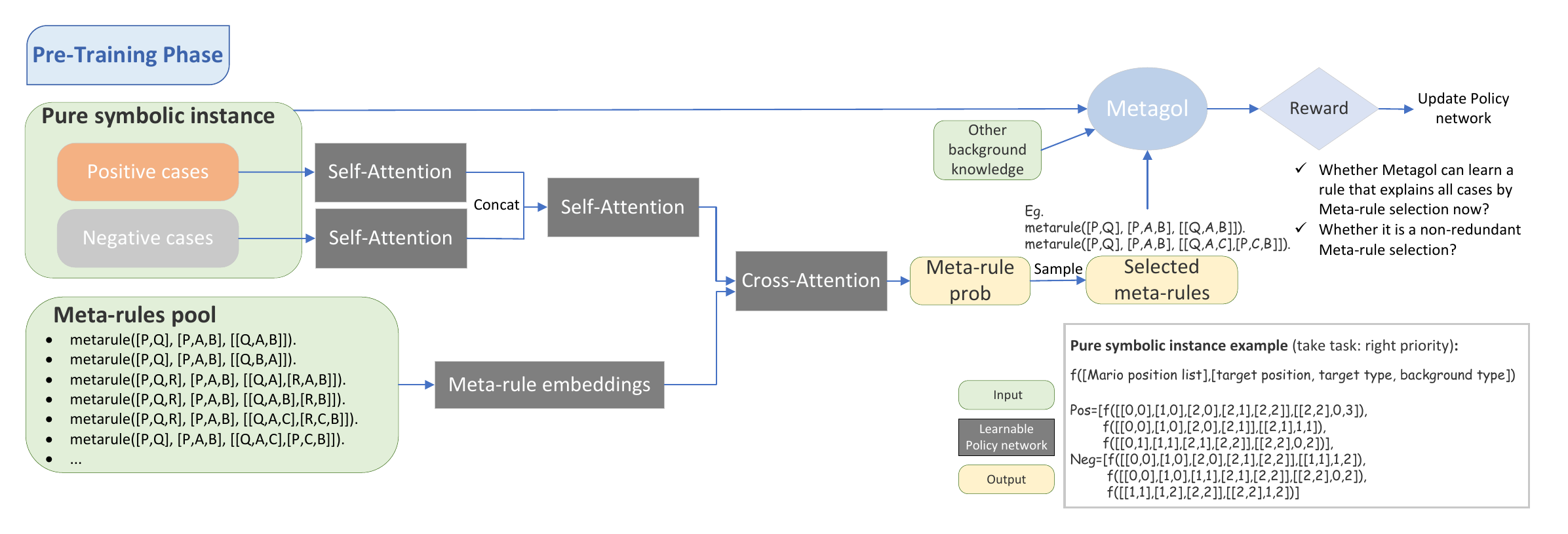}
\caption{The pre-training phase for learning the meta-rule selection policy.}
\label{fig:Pre-Training Phase}
\end{figure*}

\begin{figure*}[t]
\centering
\includegraphics[width=.9\linewidth]{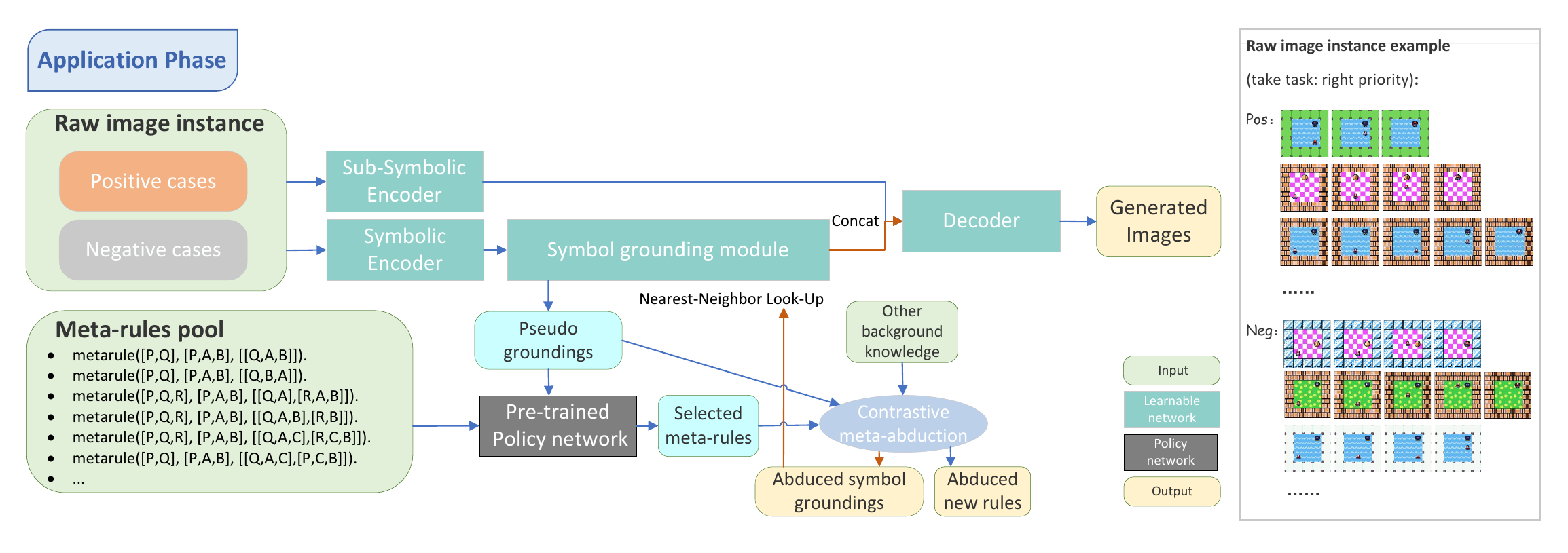}
\caption {The application phase for applying the policy in AbdGen training.}
\label{fig:Application Phase}
\end{figure*}

\section{Proposed Method}
\label{sec:method}

In this section, we introduce the pre-training strategy for learning the meta-rule selection policy and its application in \method\;to accelerate the logical programming system by selecting meta-rules from meta-rules pool. The process is divided into two main phases: the pre-training phase and the application phase, as illustrated in~\citefig{fig:Pre-Training Phase} and~\citefig{fig:Application Phase}.

The pre-training phase is illustrated in~\citefig{fig:Pre-Training Phase}, which is designed to teach the model to identify and select relevant meta-rules based on the symbol values extracted from various cases. The selection policy is designed a neural model, which employs multiple attention mechanisms to process the symbol grounding representations, and utilizes a cross-attention mechanism to generate the probabilities for meta-rule selection. Bernoulli sampling is then applied to these probabilities to select the meta-rules that will be used in the logical programming system. Since only pure symbol data is involved in pre-training, we can utilize meta-interpretative learning method (Metagol~\citep{metagol}), which is much more time-efficient than meta-abduction method in AbdGen since no symbol grounding learning is involved. The performance of the selected meta-rules is evaluated based on whether Metagol can learn a rule that explains all cases and the number of selected meta-rules. This evaluation informs the reward calculation, which is utilized to update the policy model via neural policy optimization algorithm.

The application phase is illustrated in~\citefig{fig:Application Phase}, which leverages the pre-trained model to expedite the meta-rule selection process for the training instances, which are presented as raw images, with only a small subset of their symbol groundings being labeled. The pre-trained policy takes pseudo-groundings that the neuro-symbolic visual generator predicts, and output a shrinked subset of meta-rule pool for abductive learning, hence significantly reduces the rule search space and saves the time cost of AbdGen training. 

\subsection{Model Design}

To facilitate meta-rule selection, a series of attention mechanisms~\citep{bahdanau2014neural} are employed (\citefig{fig:Pre-Training Phase}). The input to this process is a pure symbol cases bag adhering to specific task rules, including both positive and negative cases. The output is the meta-rule selection most relevant to the case patterns, sampled by meta-rule probabilities, which is calculated by the following process.

{\noindent\bf Self-attention}. Recall that each instance consists of positive and negative cases. For the set of positive cases, a self-attention mechanism is applied to their symbol embedding representations. This mechanism is designed to identify internal correlations within the positive cases, capturing important patterns and dependencies that are crucial for subsequent processing. Similarly, the negative cases undergo a self-attention process to uncover internal relationships among them. After processing both positive and negative cases individually, the resultant tensor embeddings from the self-attention mechanisms are concatenated to form a combined tensor. This combined tensor embodies the interactions between positive and negative cases. To delve deeper into these interactions, another layer of self-attention is applied to the combined tensor. This step enhances the model's understanding of the correlations between the cases, integrating information from both positive and negative cases to create a comprehensive representation.

{\noindent\bf Cross-attention}. Cross-attention mechanism, which further involves the learnable embeddings of meta-rules, aims to link the combined tensor of instance with the meta-rules to generate selection probabilities for each meta-rule. The cross-attention mechanism takes the combined tensor and meta-rule embeddings as inputs and outputs a probability for each meta-rule, incorporating information from the cases within the instance. This mechanism allows the model to assess the relevance of each meta-rule to the given cases.

{\noindent\bf Sampling}. Based on the selection probabilities obtained from the cross-attention mechanism, Bernoulli sampling is performed to select the most relevant meta-rules. The sampling strategy is crucial as it determines which meta-rules will be used in the logical reasoning process. By assigning higher probabilities to promising meta-rules, the model enhances the efficiency of the selection process, ensuring that only the most pertinent meta-rules are chosen.

\subsection{Training Method}
\label{subsec:conabd}
{\bf Policy optimization in pre-training stage.} We regard the selection of meta-rules as a decision-making process. Specifically, based on the positive and negative samples of the current task, we choose a subset of meta-rules for subsequent rule learning through a selection policy. Our insight is that when the negative and positive cases are sufficiently provided, if the meta-rule selection is inappropriate, it is even impossible to induce a rule by Metagol. Therefore, we set the reward $R$ based on whether rules can be successfully learned:
\begin{equation}
R=\left\{
\begin{aligned}
&0, & &\text{if Metagol cannot induce a rule},\\
&2^{n-n_s}, & &\text{otherwise},
\end{aligned}
\right.
\label{eq:reward}
\end{equation}
where $n$ is the total number of meta-rules in the meta-rule pool and $n_s$ represents the number of selected meta-rules. We preserve the trajectory of the decision-making process for subsequent policy updates stage. In each iteration of policy updates, we use proximal policy optimization (PPO)~\citep{schulman2017proximal} to optimize the policy. We treat a bag of instances sampled in each iterations as a trajectory and calculate the advantage function based on it, and further updates the policy network according to the clipped gradients.

\section{Experiments}
\label{sec:exp}

We conducted a series of experiments to verify the following questions: 1) Whether our meta-rule selection policy can enable AbdGen to utilize data more effectively and reduce training time compared to using all meta-rules and random meta-rules? 2) Whether our method can approach the performance level of handmade optimized meta-rules without human bias? 3) Whether our meta-rule selection policy can generalize to unseen tasks that have patterns similar to those in the training tasks? The neural and logical components of our method are implemented using PyTorch~\citep{paszke2019pytorch} and SWI-Prolog~\citep{wielemaker2003overview}, respectively. The experiments were conducted on server clusters equipped with NVIDIA RTX A5000 and 4090 GPUs. Major experimental setups are presented in the following section and more details are provided in the appendix.

\subsection{Experimental Setup}
{\bf Meta rules pool.} The meta-rules pool used in our experiments is summarized in ~\citetab{tab: meta-rule pool}. This pool includes 6 classic and commonly used logical meta-rules that the model can select to use in logical programming system. The meta-rules are designed to represent different logical patterns and relationships.

\begin{table}[t]
    \caption{Meta-rules pool}
    \centering
    \resizebox{.5\linewidth}{!}{
    \scriptsize
    \begin{tabular}{ccccc}
    \toprule
        Name & Meta-rule \\
\midrule
        Identity & $\mathtt{metarule([P,Q], [P,A,B], [[Q,A,B]]).}$ \\
        Inverse & $\mathtt{metarule([P,Q], [P,A,B], [[Q,B,A]]).}$ \\
        Precon & $\mathtt{metarule([P,Q,R], [P,A,B], [[Q,A],[R,A,B]]).}$ \\
        Postcon & $\mathtt{metarule([P,Q,R], [P,A,B], [[Q,A,B],[R,B]]).}$ \\
        Chain & $\mathtt{metarule([P,Q,R], [P,A,B], [[Q,A,C], [R,C,B]]).}$ \\
        Recursion & $\mathtt{metarule([P,Q], [P,A,B], [[Q,A,C],[P,C,B]]).}$ \\
    \bottomrule
    \end{tabular}
    }
    \label{tab: meta-rule pool}
\end{table}

{\noindent \bf Datasets and Tasks.} We conducted our experiments on two datasets: the Mario dataset~\citep{misino2022vael} and the MNIST dataset~\citep{lecun2010mnist}. Each dataset consists of a variety of tasks, each designed to test the ability of our meta-rule selection policy to find the most suitable meta-rules corresponding to the data patterns of each task. It is important to note that each task in the datasets is independent and does not influence the others. The tasks are summarized in ~\citetab{tab:Mario tasks} and ~\citetab{tab:MNIST tasks}. The tasks underlined in ~\citetab{tab:Mario tasks} are unseen during the Pre-training phase, but there are tasks with similar patterns in the train tasks.

\begin{table}
    \caption{Task description on Mario dataset.}
    \centering
    \resizebox{.9\textwidth}{!}{
    \scriptsize
    \begin{tabular}{ccccc}
    \toprule
        \makecell[c]{Task} & 
        \makecell[c]{Rule} & 
        \makecell[c]{Meta-Rule} & 
        \makecell[c]{Description} &
        \makecell[c]{Case Length}\\
    \midrule 
        \makecell[c]{Right priority \\ (up, left,
        \underline{down}) \\ } & 
        \makecell[c]{$\mathtt{f(A, B)}$ :-$\mathtt{right(A,C), f(C, B). }$ \\ 
        $\mathtt{f(A, B)}$ :-$\mathtt{f_1(A, B).}$ \\
        $\mathtt{f_1(A, B)}$ :-$\mathtt{up(A,C),f_1(C,B).}$ \\ 
        $\mathtt{f_1(A, B)}$ :-$\mathtt{terminate(A, B).}$} & 
        \makecell[c]{Identity \\ Recursion} &
        \makecell[c]{Mario follows the path \\ to the right and then up \\ to reach the target.} &
        \makecell[c]{2,3,4,5}\\
    \midrule
        \makecell[c]{Just up \\ (down, left, \underline{right})} & 
        \makecell[c]{$\mathtt{f(A, B)}$ :-$\mathtt{up(A, C),terminate(C, B).}$ \\ $\mathtt{f(A, B)}$ :-$\mathtt{up(A, C),f_1(C, B).}$ \\ 
        $\mathtt{f_1(A, B)}$ :-$\mathtt{up(A, C),terminate(C, B).}$} & 
        \makecell[c]{Chain} &
        \makecell[c]{Mario follows the path \\ just to the up \\to reach the target.}&
        \makecell[c]{2,3}\\
    \midrule
        \makecell[c]{Right one step \\ (down, left, \underline{up})} & 
        \makecell[c]{$\mathtt{f(A, B)}$ :-$\mathtt{right(A, C),terminate(C, B).}$} &
        \makecell[c]{Chain} &
        \makecell[c]{Mario moves one step \\ to the right \\to reach the target.}&
        \makecell[c]{2} \\
    \midrule
        Bomb far & 
        \makecell[c]{$\mathtt{f(A ,B)}$ :-$\mathtt{far(A, B), bomb(B)}$} & 
        \makecell[c]{Postcon} &
        \makecell[c]{If target is a bomb,\\ Mario goes further.} &
        \makecell[c]{2}\\
    \midrule
        Flower &
        \makecell[c]{$\mathtt{f(A,B)}$ :-$\mathtt{right(A,C),f(C,B).}$ \\ 
        $\mathtt{f(A,B)}$ :-$\mathtt{f_1(A,B),flower(B).}$ \\
        $\mathtt{f_1(A,B)}$ :-$\mathtt{up(A,C),f_1(C,B).}$ \\ 
        $\mathtt{f_1(A,B)}$ :-$\mathtt{terminate(A,B),flower(B).}$} & 
        \makecell[c]{Postcon\\Chain} &
        \makecell[c]{If background is flower,\\Mario follows the path \\ to the right and then up \\ to reach the target.}  &
        \makecell[c]{2,3,4,5}\\
    \midrule
        Chess jump & 
        \makecell[c]{$\mathtt{f(A ,B)}$ :-$\mathtt{terminate(A,B),chess(B).}$ \\
        $\mathtt{f(A,B)}$ :-$\mathtt{jump(A,C),f(C,B).}$} & 
        \makecell[c]{Postcon\\Chain} &
        \makecell[c]{If background is chess,\\Mario jumps Diagonally \\ to reach the target.}  &
        \makecell[c]{2,3,4,5}\\
    \bottomrule
    \end{tabular}
    }
    \label{tab:Mario tasks}
\end{table}

\begin{table}[t]
    \caption{Task description on MNIST dataset.}
    \centering
    \resizebox{.9\linewidth}{!}{
    \scriptsize
    \begin{tabular}{ccccc}
    \toprule
        \makecell[c]{Task} & 
        \makecell[c]{Rule} & 
        \makecell[c]{Meta-Rule} &
        \makecell[c]{Description} &
        \makecell[c]{Case Length}\\ 
    \midrule
        \makecell[c]{Add priority\\(multi)} & \makecell[c]{$\mathtt{f(A,tB)}$:-$\mathtt{add(A,C),f(C,B).}$ \\ $\mathtt{f(A, B)}$:-$\mathtt{f_1(A,B).}$ \\ $\mathtt{f_1(A,B)}$:-$\mathtt{multi(A,C),f_1(C,B).}$ \\ $\mathtt{f_1(A,B)}$:-$\mathtt{eq(A,B).}$} & 
        \makecell[c]{Identity \\ Recursion} &
        \makecell[c]{Add first then multi on \\ a list of uncertain length.\\Result is given last}  &
        \makecell[c]{3,4,5}\\
    \midrule
        \makecell[c]{Cumulative sum \\ (product)} & 
        \makecell[c]{$\mathtt{f(A,B)}$:-$\mathtt{add(A,C),f(C,B).}$ \\ $\mathtt{f(A,B)}$:-$\mathtt{eq(A,B).}$} & 
        \makecell[c]{Identity \\ Recursion} &
        \makecell[c]{Cumulative sum on \\ a list of uncertain length.\\Result is given last}  &
        \makecell[c]{3,4,5}\\
    \midrule
        \makecell[c]{Reverse \\ cumulative sum \\ (product)} & \makecell[c]{$\mathtt{f(A,B)}$:-$\mathtt{f_1(B,A).}$ \\ $\mathtt{f_1(A,B)}$:-$\mathtt{add(A,C),f_1(C,B).}$ \\ $\mathtt{f_1(A,B}$:-$\mathtt{f_2(B,A).}$ \\ $\mathtt{f_2(A,B)}$:-$\mathtt{eq(A,B).}$} & 
        \makecell[c]{Inverse \\ Recursion} &
        \makecell[c]{Cumulative sum on \\ a list of uncertain length. \\ Result is given first.}  &
        \makecell[c]{3,4,5}\\
    \midrule
        \makecell[c]{Increasing \\ sequence} & 
        \makecell[c]{$\mathtt{f(A,B)}$:-$\mathtt{less(A,C),f(C,B).}$ \\ $\mathtt{f(A,B)}$:-$\mathtt{less(A,B),zero(B).}$} & 
        \makecell[c]{Postcon \\ Recursion}  &
        \makecell[c]{If the last number is zero,\\ the previous list is\\ a increasing sequence.}  &
        \makecell[c]{3,4,5}\\
    \midrule
        \makecell[c]{Decreasing \\ sequence} & 
        \makecell[c]{$\mathtt{f(A,B)}$:-$\mathtt{f_1(B,A).}$ \\ $\mathtt{f_1(A,B)}$:-$\mathtt{more(A,C),f_1(C,B).}$ \\ $\mathtt{f_1(A,B)}$:-$\mathtt{more(A,B),zero(B).}$} & 
        \makecell[c]{Inverse \\ Postcon \\ Recursion}  &
        \makecell[c]{If the first number is zero, \\the following list is \\a decreasing sequence.}  &
        \makecell[c]{3,4,5}\\
    \bottomrule
    \end{tabular}
    }
    \label{tab:MNIST tasks}
\end{table}

{\noindent \bf Comparison methods.} To evaluate the effectiveness of our meta-rule selection policy, we compared it with the following baseline methods:

\begin{itemize}
\item[$\bullet$] Handmade optimized meta-rule: Manually crafted optimal meta-rules tailored for each task.
\item[$\bullet$] Random Meta-rule: Randomly selected two meta-rules from the meta-rule pool for AbdGen training.
\item[$\bullet$] All meta-rule: Utilizing all meta-rules in the meta-rule pool for training.
\end{itemize}

{\noindent \bf Performance metrics.} We used several performance metrics to evaluate our meta-rule selection model's effectiveness and efficiency applied to AbdGen:

\begin{itemize}
\item[$\bullet$] Symbol grounding accuracy: The symbol grounding accuracy of the model as training iterations progress.
\item[$\bullet$] Training time: The time to reach a certain training iterations. 
\item[$\bullet$] Rule learning performance: This metric includes the rate of {\it success}, {\it timeout} (proportion of the trials where the system times out, set at a threshold of 30 seconds), and {\it other error} (proportion of the trials where the system returns no suitable rule) for rule learning.
\item[$\bullet$] Rule based Generation Quality: The quality of the generated outputs, measured by the adherence of generated images to the specified logical rules and the visual consistency of these images. Due to space limitation, the illustration of generation results are provided in the appendix.

\end{itemize}

\subsection{Results}

{\bf Pre-Training Phase Findings.} Our experiments yielded two significant findings during the pre-training phase:

\begin{itemize}
\item[1)] Discovery of fewer meta-rules: Our meta-rule selection model was able to find solutions that required fewer meta-rules than those initially preset. Specifically, for the Mario Just Up and MNIST Reverse Cumulative Sum tasks, we found that fewer meta-rules could explain all cases. ~\citetab{tab:discovery} shows that the discovered rules used fewer meta-rules compared to the preset rules. In the Mario Just Up task, the reduction in meta-rules is attributed to the constraint of the 3x3 Mario grid map, which allows for a simpler explanation. For the MNIST Reverse Cumulative Sum task, the fewer meta-rules are due to the neural networsk's reward-driven approach, which can discover higher-reward explanations that may be outside the linear, sequential thinking typically employed by humans. Namely, the rules generated with fewer meta-rules may require a slight shift in thinking for humans to understand. This indicates that the model can optimize the meta-rule selection process, finding more efficient logical meta-rules to solve the tasks.
\item[2)] Generalization to Unseen Tasks: We observed an unexpected but promising behavior where the meta-rule selection probabilities for unseen tasks with patterns similar to the training tasks also adjusted in the correct direction as training progressed. ~\citefig{fig:prob_changing} shows the heatmaps of the changing probabilities for different meta-rules being selected for both exemplified training tasks and unseen tasks. These results suggest that our model can generalize well to new pattern-like tasks, effectively transferring knowledge from the training tasks to unseen tasks.
\end{itemize}

\begin{table}[t]
    \caption{Discovery of fewer meta-rules}
    \centering
    \resizebox{.9\linewidth}{!}{
    \scriptsize
    \begin{tabular}{ccccc}
    \toprule
        \makecell[c]{Task} & 
        \makecell[c]{Preset Rule} & 
        \makecell[c]{Preset Meta-Rule} &
        \makecell[c]{Discovered Rule} &
        \makecell[c]{Fewer Meta-Rule} \\ 
\midrule
     \makecell[c]{Mario \\Just up} & 
     \makecell[c]{$\mathtt{f(A,B)}$:-$\mathtt{up(A,C),}\mathtt{f(C,B).}$ \\ $\mathtt{f(A,B)}$:-$\mathtt{terminate(A,B).}$} & 
     \makecell[c]{Identity \\ Recursion}  &
     \makecell[c]{$\mathtt{f(A, B)}$ :-$\mathtt{up(A, C),terminate(C, B).}$ \\ 
     $\mathtt{f(A, B)}$ :-$\mathtt{up(A, C),f_1(C, B).}$ \\ 
     $\mathtt{f_1(A, B)}$ :-$\mathtt{up(A, C),terminate(C, B).}$} &
     \makecell[c]{Chain} \\
     \midrule
     \makecell[c]{MNIST \\ Reverse cumulative Sum} & 
     \makecell[c]{$\mathtt{f(A,B)}$:-$\mathtt{f_1(B,A).}$ \\ $\mathtt{f_1(A,B)}$:-$\mathtt{add(A,B), f_1(C, B).}$\\
     $\mathtt{f_1(A,B)}$:-$\mathtt{eq(A,B).}$}& 
     \makecell[c]{Identity\\Inverse \\ Recursion}  &
     \makecell[c]{$\mathtt{f(A,B)}$:-$\mathtt{f_1(B,A).}$ \\ $\mathtt{f_1(A,B)}$:-$\mathtt{add(A,C),f_1(C,B).}$ \\ $\mathtt{f_1(A,B)}$:-$\mathtt{f_2(B,A).}$ \\ $\mathtt{f_2(A,B)}$:-$\mathtt{eq(A,B).}$} & 
     \makecell[c]{Inverse \\ Recursion} \\
    \bottomrule
    \end{tabular}
    }
    \label{tab:discovery}
\end{table}

\begin{figure*}[t]
\centering
\includegraphics[width=.9\linewidth]{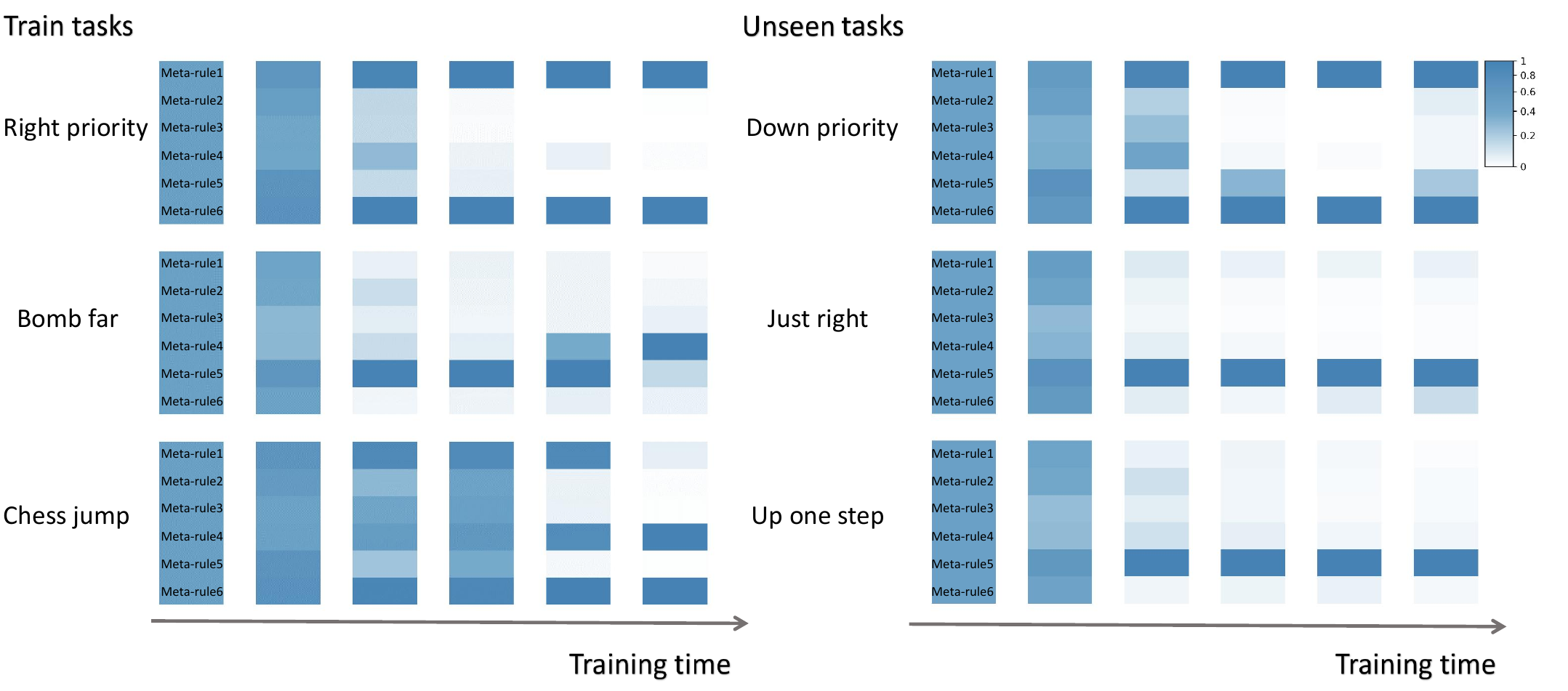}
\caption {The changing probabilities of different meta-rules being selected for different tasks during pre-training.}
\label{fig:prob_changing}
\end{figure*}

\begin{figure*}[t]
    \centering
    \begin{subfigure}[t]{.24\textwidth}
      \includegraphics[width=\textwidth]{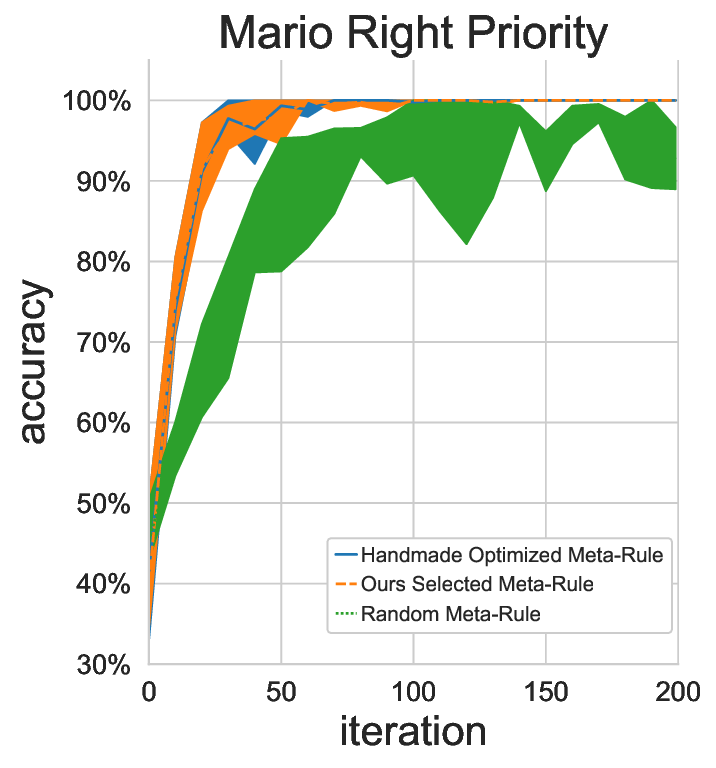}
    \end{subfigure}
    \begin{subfigure}[t]{.24\textwidth}
        \includegraphics[width=\textwidth]{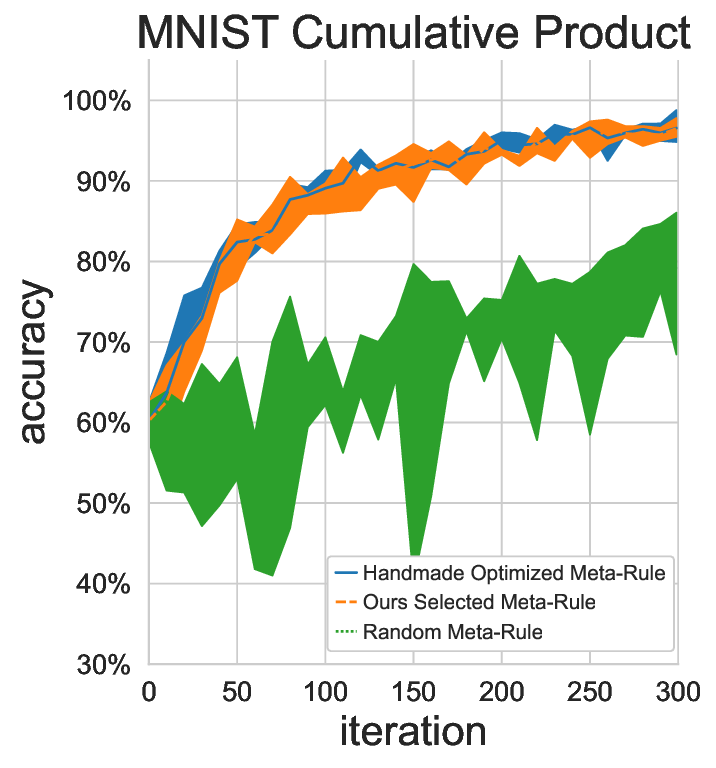}
    \end{subfigure}
    \begin{subfigure}[t]{.24\textwidth}
      \includegraphics[width=\textwidth]{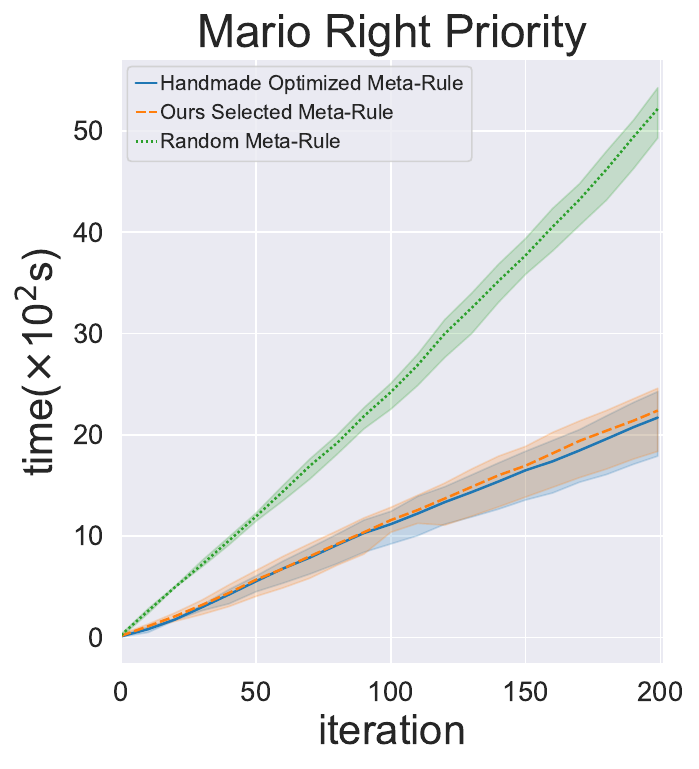}
    \end{subfigure}
    \begin{subfigure}[t]{.24\textwidth}
      \includegraphics[width=\textwidth]{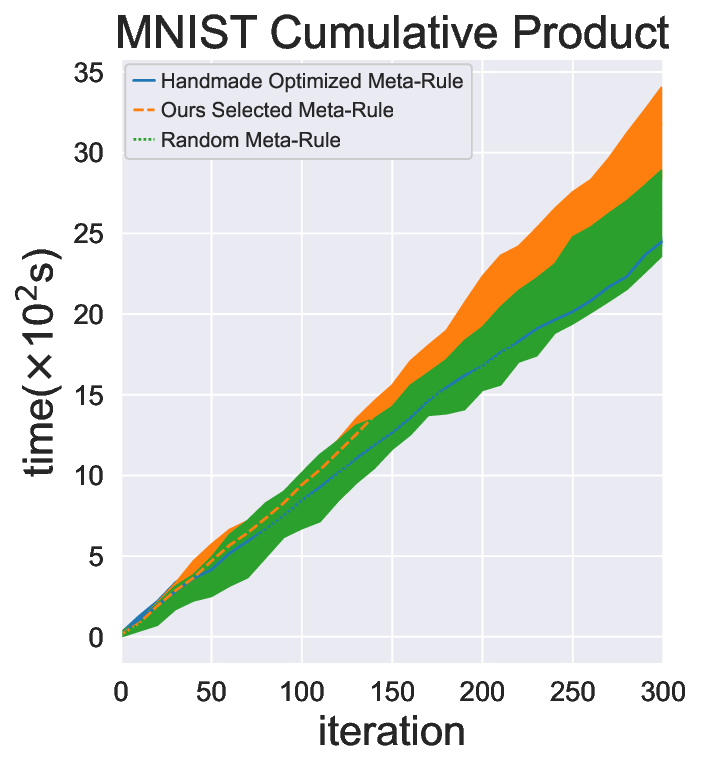}
    \end{subfigure}
    \caption{{\bf Performance on train tasks.} Comparisons on train tasks Mario {\it Right priority} and MNIST {\it Cumulative product} of our meta-rule selection model, hand-made optimized meta-rules and random meta-rules on grounding accuracy ($\uparrow$) and training time($\uparrow$)  over AbdGen iterations.}
    \label{fig:train task results}
\end{figure*}

\begin{figure*}[t]
    \centering
    \begin{subfigure}[t]{.24\linewidth}
    \includegraphics[width=\linewidth]{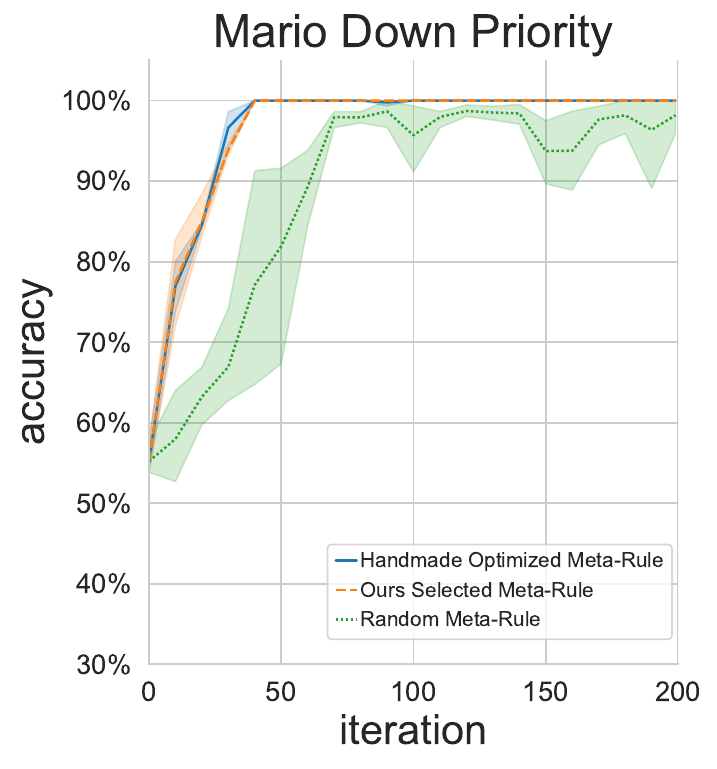}
    \end{subfigure}    
        \begin{subfigure}[t]{.24\linewidth}
        \includegraphics[width=\linewidth]{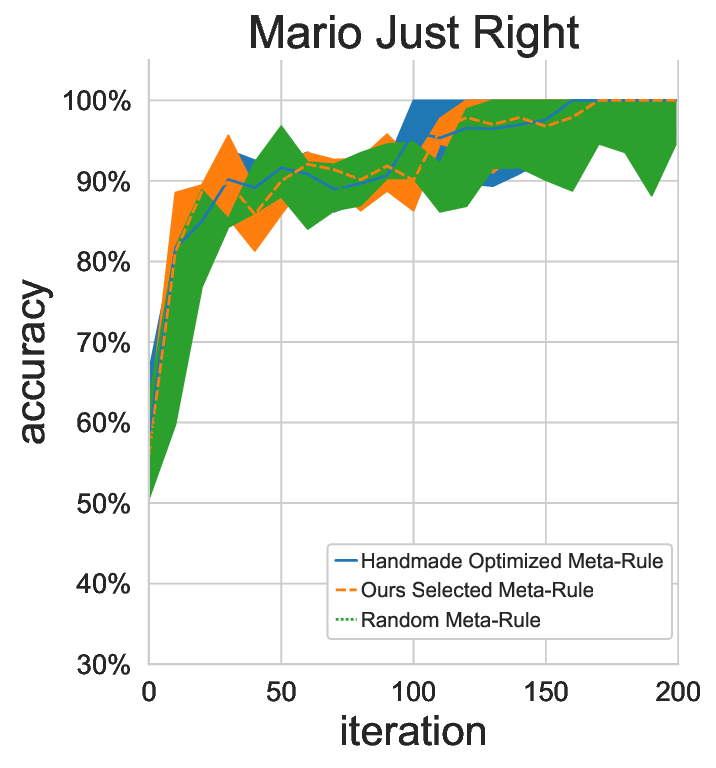}
    \end{subfigure}
    \begin{subfigure}[t]{.24\linewidth}
    \includegraphics[width=\linewidth]{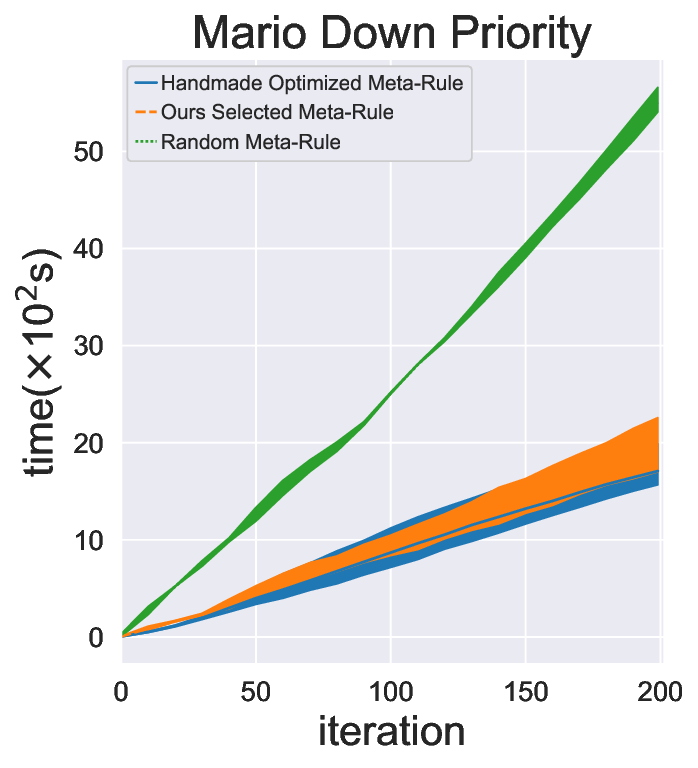}
    \end{subfigure}
    \begin{subfigure}[t]{.24\linewidth}
        \includegraphics[width=\linewidth]{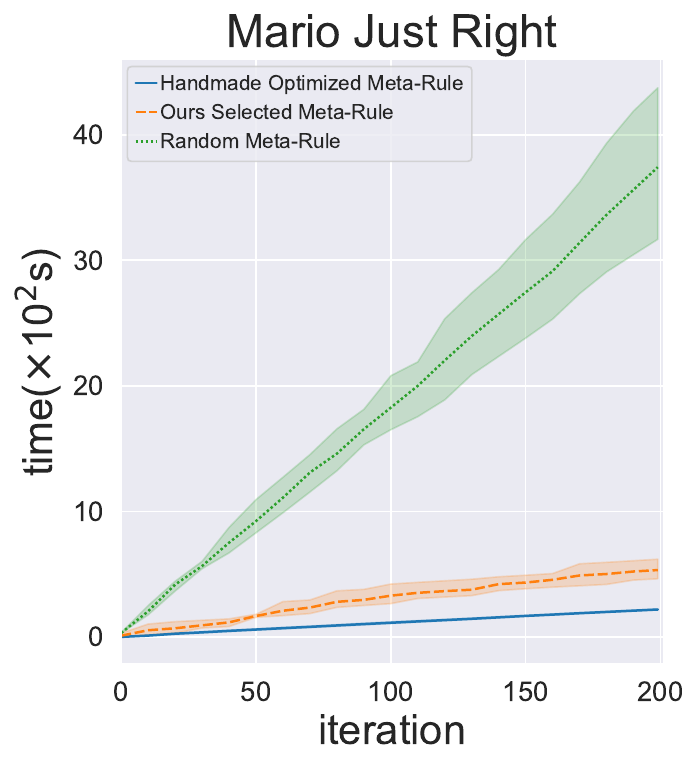}
    \end{subfigure}  
    \caption{{\bf Performance on unseen tasks.} Comparisons on unseen tasks Mario {\it Down priority} and Mario {\it Just Right} of our meta-rule selection model, hand-made optimized meta-rules and random meta-rules on grounding accuracy ($\uparrow$) and training time($\uparrow$)  over AbdGen iterations.}
    \label{fig:test task results}
\end{figure*}

\begin{table}[t]
    \caption{Performance comparison.}
    \centering
    \resizebox{.9\linewidth}{!}{
    \scriptsize
    \begin{tabular}{ccccc}
    \toprule
        Task & Method & Timeout Rate & Other Error Rate& Success Rate\\
    \midrule
        \multirow{4}{*}{\shortstack{Mario \\ Right Priority}} & Handmade optimized meta-rule & 0.005$\pm$0.001 & 0.036$\pm$0.012 & 0.959$\pm$0.012 \\
        & Ours selected meta-rule & 0.006$\pm$0.002 & 0.120$\pm$0.003 & 0.874$\pm$0.002 \\
        & Random meta-rule & 0.147$\pm$0.007 & 0.692$\pm$0.005 & 0.161$\pm$0.003 \\
        & All meta-rule & 1.000$\pm$0.000 & 0.000$\pm$0.000 & 0.000$\pm$0.000 \\
    \midrule
        \multirow{4}{*}{\shortstack{MNIST \\Cumulative Product}} & Handmade optimized meta-rule & 0.045$\pm$0.011 & 0.003$\pm$0.002 & 0.952$\pm$0.009 \\
        & Ours selected meta-rule & 0.111$\pm$0.009 & 0.077$\pm$0.011 & 0.812$\pm$0.007 \\
        & Random meta-rule & 0.067$\pm$0.002 & 0.760$\pm$0.014 & 0.173$\pm$0.013 \\
        & All meta-rule & 1.000$\pm$0.000 & 0.000$\pm$0.000 & 0.000$\pm$0.000 \\
    \bottomrule
    \end{tabular}
    }
    \label{tab:result table}
\end{table}

{\noindent\bf Symbol Grounding accuracy and Training Time over AbdGen iterations.} The expriments of application phase are conducted on the Mario {\it Right priority} and MNIST {\it Cumulative product} tasks. We compare the performance of our meta-rule selection model with handmade optimized meta-rules and random meta-rules. The objective is to evaluate how effectively our model can select most correct meta-rules and then boost the symbol grounding and the time efficiency over AbdGen training iterations.

~\citefig{fig:train task results} presents the results of our symbol grounding experiments. The left two subfigures illustrate the symbol grounding accuracy over iterations for the tasks Mario {\it Right priority} and MNIST {\it Cumulative product}. It is evident that our selected meta-rules achieve comparable accuracy to handmade optimized meta-rules and significantly outperform random meta-rules. The right two subfigures depict the training time over iterations for the same tasks. For Mario {\it Right priority} task, our selected meta-rules demonstrate a significant reduction in training time compared to the use of random meta-rules. The peorformance is close to that of handmade optimized meta-rules, demonstrating the accuracy of our pre-trained model in selecting meta-rules even when there is significant noise in the symbol values. Conversely, for MNIST {\it Cumulative product} task, the training time for random meta-rules is shorter. This task has fewer predicates prepared (5 predicates: $\mathtt{add, multi, eq, head, empty}$), resulting in a smaller search space. Many incorrect meta-rule combinations quickly lead to no rule being learned, which results in a rapid output but zero data utilization.

~\citefig{fig:test task results} illustrates the performance of our pre-trained meta-rule selection model on unseen tasks with similar patterns to the training tasks, which is a discovery in the pre-training phase. These results demonstrate that our model can effectively generalize to new tasks it has not encountered during pre-training, maintaining a high level of selecting correct meta-rules. For Mario {\it Just Right} task by randomly selecting meta-rules, as long as selecting {\it Chain} as one of the meta-rules can derive the rule. Therefore, the difference in the grounding accuracy curve is not as pronounced as in the other tasks. However, the time consumption for random meta-rule selection is significantly higher for redundant or wrong meta-rules. Furthermore, the results of all meta-rules baseline are not shown in~\citefig{fig:train task results} and~\citefig{fig:test task results} because each training iteration results in a timeout.

\citetab{tab:result table} provides a detailed performance comparison across different methods, including the timeout rate, other error rate, and success rate of logical programming system with differently sampled meta-rule inputs. Overall, these results demonstrate that our pre-trained model can select meta-rules without human intervention, achieving a data utilization level close to that of manually chosen meta-rules. For the rates of MNIST {\it Cumulative product} task by random meta-rule, We can interpret that as only when meta-rule {\it Inverse} and meta-rule {\it Recursion} are selected simultaneously does the search fall into a loop, causing a timeout. When the selected meta-rules are {\it Identity} and {\it Recursion}, the data can be utilized for sure. Additionally, if the sampled 2 positive cases both have a length of only 3, selecting {\it Chain} as one of the meta-rules can also explain these cases, which is similar to the concept illustrated in \citetab{tab:discovery}.


{\noindent \bf Robustness to unseen grounding errors.} An additional interesting observation from the experiments is the robustness of the selection policy to unseen grounding errors. During pre-training phase, the symbol groundings of the data are fully labeled, thus the selection policy can be trained under ground-truth groundings. However, during application phase, especially at the early period of AbdGen training, the pseudo groundings are usually inaccurate and unseen during pre-training. A small set of labeled data is introduced in the application phase, which avoids the AbdGen model to make too large grounding errors. Interestingly, it is observed that ensuring the initial grounding error to be below 50\% is generally sufficient to achieve the desirable performance. In our view, this phenomenon verifies the memorization ability of attention mechanism~\citep{ramsauer2020hopfield}, which leads to the recovery from deviations of local minima, and the relative stability of symbolic patterns, thus is worth further researches to investigate. 

\section{Conclusion}
In this paper, we propose the method of pre-training the meta-rule selection policy for boosting the training efficiency of visual generative abductive learning. The method builds upon the embedding mechanism 
 as well as the disentanglement of symbolic and sub-symbolic information of the AbdGen model~\citep{peng2023generating}, which can utilize pure symbolic data for pre-training, hence is of low additional time cost. The experimental results show that the proposed approach the policy effectively reduces the time cost of AbdGen training, meanwhile it can also deal with unseen learning tasks and robust to unseen grounding errors. 

\section{Acknowledgement}
This work is supported by National Key R\&D Program of China (2022ZD0114804) and National Natural Science Foundation of China (62206245).
\bibliographystyle{named}
\bibliography{nesy}
\clearpage
\appendix
\section{Detailed Experimental Setups}

\subsection{Data configurations}

\begin{itemize}
\item[$\bullet$] {\bf Mario Dataset.} For the Mario dataset, we consider a 3x3 grid map with target types divided into coin and bomb. The background types include sea, flower, chessboard. The frame types are categorized as brick1, brick2, brick3, green panel, white panel, glass, and concrete. This results in a total of 3402 images, corresponding to 3402 sets of symbolic data. For each task, an image (symbol) case contains between 2 to 5 images (symbol data sets), as detailed in Table 2 of the main paper.
\begin{itemize}
\item[-]{\bf Pre-training phase.} For the pre-training process, we use the pure symbolic instance of all tasks. Due to the different settings of each task, the number of positive cases that can be generated is limited. For the {\it Priority} task, 1134 positive cases can be formed. For the {\it Just} and {\it Flower} tasks, 378 positive cases each can be formed. The {\it One Step} task can form 252 positive cases. The {\it Sea} task can form 126 positive cases, and the {\it Chess jump} task can form 3640 positive cases. Negative cases, in theory, can be any combination that is not a positive case. However, to ensure that the negative cases are informative for rule learning, we increased the proportion of meaningful negative cases for each task. We uniformly generated 3000 negative cases for each task. In this phase, we do not specify the number of positive and negative cases in each instance. Instead, we set a maximum allowable number of 20 positive and 50 negative cases per instance that the model can handle, aiming for the application phase to adapt to varying numbers of cases, ensuring flexibility and robustness across different scenarios. 
\item[-]{\bf Application phase.} For the application phase, we use the raw image instance. For the {\it Right (Down) priority} task, there are 1134 positive cases composed of 1218 different images and 3000 negative cases composed of 2552 (2553) different images. For the {\it Just right task}, there are 378 positive cases composed of 378 different images and 3000 negative cases composed of 2776 different images. Each instance contains 5 positive cases and 20 negative cases. For each task, we initially use 1\% of the total number of images, approximately 20-30 images, for few-shot supervised training. This is done to achieve around 50\% accuracy which is an acceptable error margin for the meta-rule selection model, to activate the meta-rule selection model.
\end{itemize}
\item[$\bullet$] {\bf MNIST Dataset.} For the MNIST dataset, we start with approximately 6000 original images for each digit from 0 to 9. The task data for subsequent tasks are sampled from these original images. Each image (symbol) case contains 3 to 5 images (symbol data sets).
\begin{itemize}
\item[-]{\bf Pre-training phase.} For the pre-training process, we use the symbol cases from all tasks, similar to the Mario dataset. For each task, we sample 3000 positive cases and 3000 negative cases. Similarly to the Mario dataset, we apply informative restrictions on the generation of negative cases and set a maximum allowable number of 10 positive and 20 negative cases per instance.
\item[-]{\bf Application phase.} For the application phase, we use raw image instances. For the Cumulative Product task, we re-sample 3000 positive and 3000 negative grounding symbol cases and extract the corresponding raw data to form the raw image instance dataset. The positive cases are composed of 10,887 different images, while the negative cases are composed of 8,892 different images. Each instance contains 2 positive cases and 1 negative cases. We initially use 0.1\% of the total number of images, 20 images, for few-shot supervised training. This is done to achieve around 60\% accuracy to activate the meta-rule selection model.
\end{itemize}
\end{itemize}

\subsection{Implementation details}

\begin{itemize}
\item[$\bullet$] {\bf Embedding} The positive and negative cases are directly encoded according to their symbol grounding and then expanded to the maximum supported number of positive and negative examples. 
The dimension of the meta-rule  embedding is set to 16.

\item[$\bullet$] {\bf Attention Mechanisms} The network uses Hopfield attention layers~\citep{ramsauer2020hopfield} 
for self-attention and cross-attention with padding mask on and 8 parallel attention heads.

\item[$\bullet$] {\bf Probability Prediction} A one-layer full connect network is used to predicts the probability of selecting each meta-rule with Sigmoid  activations and clamped to ensure values are within a valid range.
\end{itemize}

\begin{figure*}[t]
\centering
\includegraphics[width=.9\linewidth]{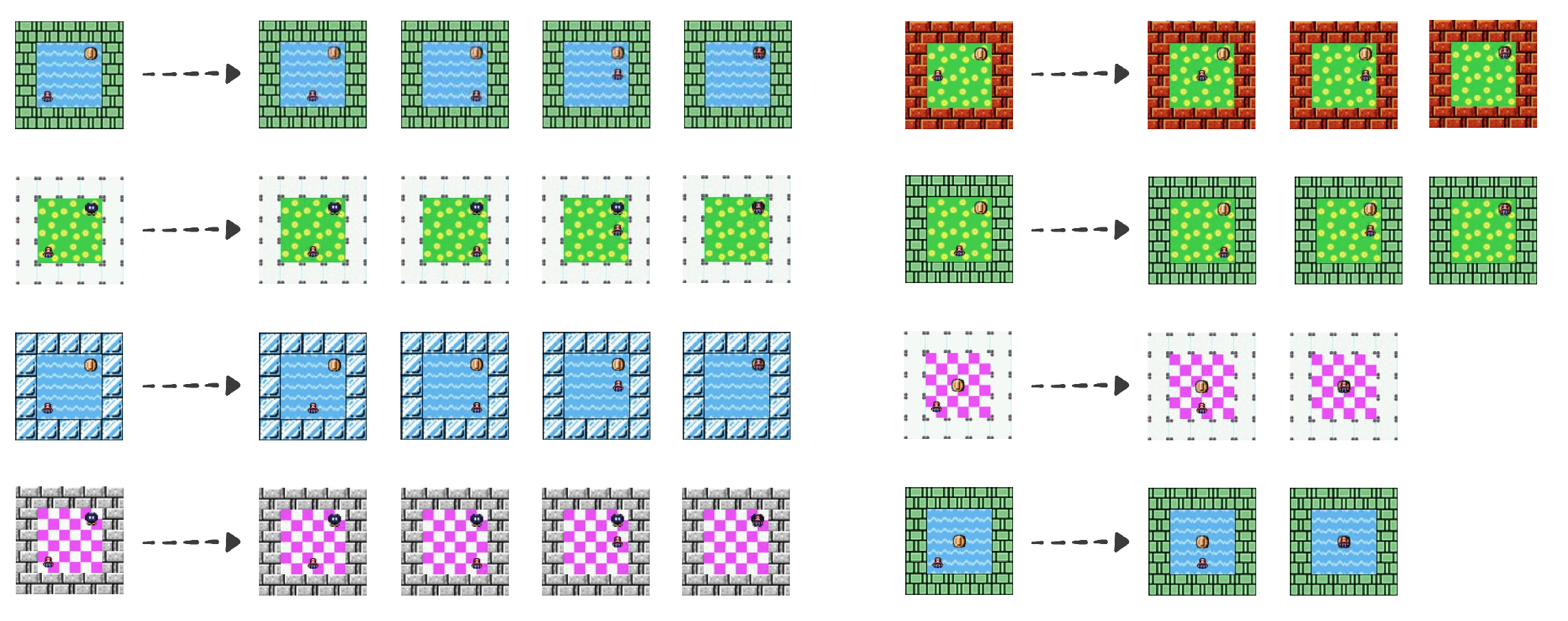}
\caption{{\bf Rule based generation on Mario.} Generation results of AbdGen on Mario {\it Right priority} task given the input image, AbdGen generates a sequence of images based on the learned correct rules.}
\label{fig:mario_gen}
\end{figure*}

\begin{figure*}[t]
\centering
\includegraphics[width=.9\linewidth]{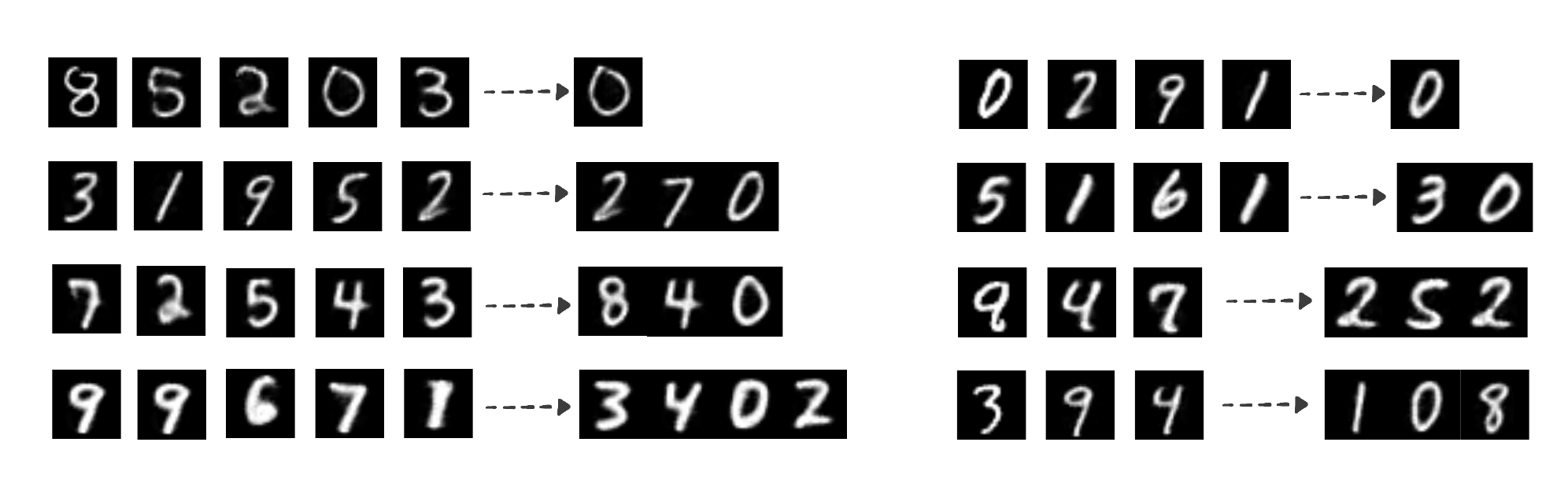}
\caption{{\bf Rule based generation on MNIST.} Generation results of AbdGen on MNIST {\it Cumulative product} task given the input images, AbdGen generates a sequence of images that represent the answer based on the learned correct rules.}
\label{fig:mnist_gen}
\end{figure*}
The pre-training process is conducted for 400 iterations for Mario dataset and 200 iterations for MNIST dataset . We utilize Adam optimizer with a learning rate of 0.00001 for both two datasets. 

For AbdGen process, the pre-training phase is conducted for 500 iterations, and the application phase is conducted for 30000 iterations. We utilize Adam optimizer~\citep{kingma2014adam} with a learning rate of 0.0001 for Mario dataset and with a learning rate of 0.001 for at the first 10000 iterations and 0.0001 at the last 20000 iterations for MNIST dataset. It should be noted that the abduction accuracy converges within 300 iterations. Therefore, Figure 5 and Figure 6 in the main paper only display the results for the first 200 or 300 iterations.

\section{Generation results}
\label{sec:exp_more}
Our generation results are shown in~\citefig{fig:mario_gen} and~\citefig{fig:mnist_gen}, which demonstrate that the AbdGen model with meta-rule selection can successfully learn the correct rules to explain all cases and generates images based on these learned rules with high quality and consistent style.
\end{document}